%% file: iclr2026_conference.tex
\documentclass{article} 
\usepackage{iclr2026_conference,times}

\input{math_commands.tex}

\usepackage{amsmath}       
\usepackage{amssymb}       
\usepackage{amsthm}        

\newtheorem{definition}{Definition}[section]

\usepackage{booktabs}      
\usepackage{array}         
\usepackage{multirow}      
\usepackage{multicol}      
\definecolor{deeppurple}{rgb}{0.36, 0.18, 0.56}
\usepackage{graphicx}      
\usepackage{subcaption}    
\usepackage{float}         

\usepackage{enumitem}      

\usepackage{xcolor}        
\usepackage{caption}       

\usepackage{hyperref}      
\usepackage{url}
\makeatletter
\long\def\blfootnote#1{\gdef\@thefnmark{*}\@footnotetext{#1}}
\makeatother

\title{Train Smarter, Not Longer: Memorization-Guided Data Reuse for Efficient LLM Training}

\author{\text{Jingwei Zuo*, Cong Zeng, Ilyas Chahed, Maksim Velikanov,} \\
\text{Dhia Eddine Rhaiem, Pasquale Balsebre, Abhay Kumar, Younes Belkada, Hakim Hacid} \\ \\
Technology Innovation Institute \\
Abu Dhabi, UAE
}

\iclrfinalcopy 
\begin{document}

\maketitle

\blfootnote{Correspondence to: jingwei.zuo@tii.ae}

\begin{abstract}
The training paradigm of large language models has shifted from traditional one-pass training to multi-epoch training, as reasonable reuse of limited high-quality data can improve both model performance and sample efficiency. Meanwhile, excessive repetition introduces the risk of overfitting and diminishing returns. Determining when and how to reuse data effectively thus emerges as a natural but under-explored question. Through a novel observation of model's \textit{Memorization Window} signals derived from loss retention dynamics and downstream evaluation scores, we propose \textit{Memorization-guided Data Reuse}, a training paradigm that adaptively determines \emph{when} and \emph{how} data should be reused, enabling principled decisions on the number of training epochs and the scheduling of data replays. Our preliminary experiments reveal a consistent memorization-driven regime: performance continues to improve with repetition far beyond current practice (e.g., the commonly cited four-epoch limit). 
%
%
While a full scheduler remains future work, these insights provide a foundation for memorization-aware training schedules, helping to determine reuse budgets and move toward training LLMs \textit{smarter rather than longer} with limited high-quality data.

\end{abstract}

\section{Introduction}

\begin{figure}[H]
\centering
\begin{subfigure}[t]{0.49\linewidth}
    \includegraphics[width=\linewidth]{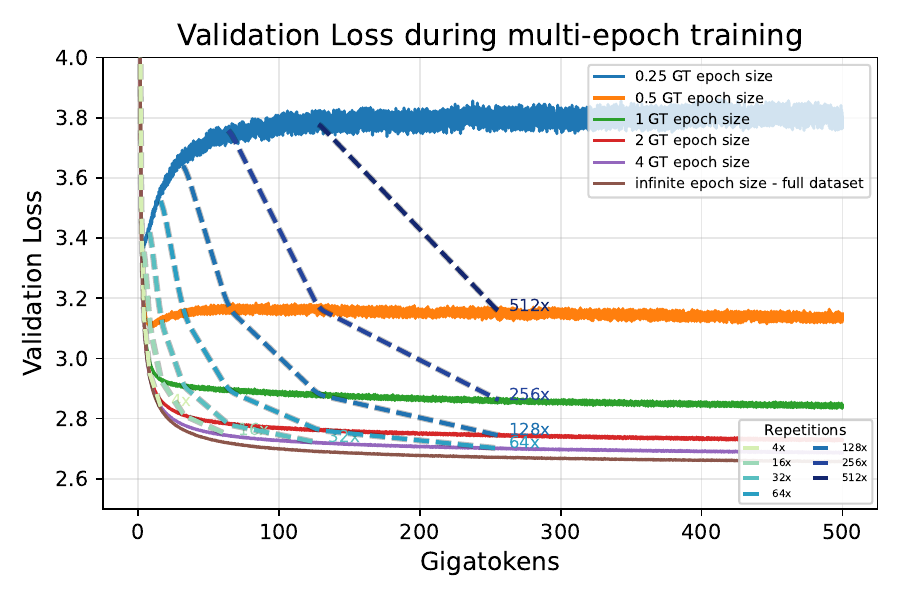}
    \caption{Epoch size effect on validation loss}
    \label{fig:epoch_size_val_loss}
\end{subfigure}
\hfill
\begin{subfigure}[t]{0.49\linewidth}
    \includegraphics[width=\linewidth]{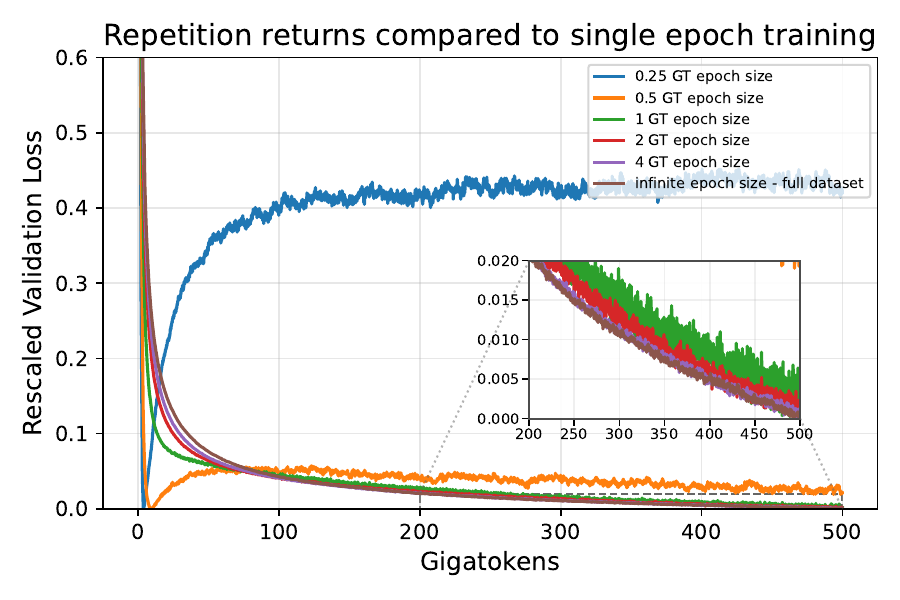}
    \caption{Marginal Returns vs. Fresh Data}
    \label{fig:returns_fresh_data}
\end{subfigure}
\caption{Epoch size determines the value of repetition. Under \textit{memorization-aware} data scheduler (100M Transformer):
(a) Validation loss for varying epoch sizes $N$. Small $N$ leads to rapid overfitting, while larger $N$ tracks the infinite-data baseline.
(b) Returns from repetition vs. fresh data. When spacing is sufficient (e.g., \textcolor{deeppurple}{$N=4.0$GT}), the loss decreases at the same rate as the full dataset (\textcolor{brown}{brown}), implying that re-learning forgotten data yields the same marginal gain as injecting new data}
\label{fig:intro_memorization}
\end{figure}
Despite most LLMs were trained with a single epoch prior to 2024, recent advances in large language model (LLM) training rely heavily on multi-epoch training, in which training data are reused across epochs to improve model performance. This practice is driven by the limited availability of high-quality (HQ) data as LLMs continue to scale. Many works~\citep{taylor2022galactica, lozhkovsmollm2, olmo2025olmo,hao2025reformulation} have explored repetition strategies, typically including the reuse of high-quality (HQ) data uniformly across epochs. 

Since prior work has explored the effect of data repetition on training~\citep{muennighoff2023scaling}, showing that \textit{training with up to four epochs of repeated data yields negligible changes to loss while further repetition leads to diminishing returns}, most existing repetition strategies adopt a relatively conservative approach. However, we argue that they either rely on fixed heuristics for data reuse (e.g., globally shuffling multi-epoch data) or ignore the dynamics of model memorization. We illustrate this dynamic in Figure~\ref{fig:epoch_size_val_loss} by varying different epoch size, the onset of overfitting is not fixed but shifts according to the epoch size $N$, effectively delaying the degradation point. 

This leads to open questions about \textit{given limited HQ data, how to optimally schedule or reuse them during training}. In practice, training with limited HQ data presents a trade-off: increasing the number of epochs on the HQ data risks overfitting through excessive repetition, while increasing HQ proportion with fixed epochs constrains the total training token budget. Addressing this trade-off requires a principled understanding of how memorization develops during training and how it interacts with data reuse.
Regarding the reuse of limited HQ data, Figure~\ref{fig:returns_fresh_data} reveals a striking finding: under \textit{memorization-aware} data scheduling, data repetition becomes mathematically equivalent to training on fresh data in terms of learning efficiency. When the spacing between repetitions is sufficient (e.g., $N=4.0$\,GT\footnote{Gigatokens (GT) correspond to billion tokens (B); the two notations are used interchangeably in the paper}), the validation loss decreases at the same rate as a model trained on fresh, unique data. This implies that while the absolute performance remains bounded by the limited diversity of the finite subset (a diversity gap), the marginal utility of a repeated token is fully restored once the model has \textit{forgotten} it, i.e., the repetition provides learning signal as if the sample were new. 

In this work, we provide preliminary empirical evidence of a \textit{Memorization Window} in LLM training and its interaction with training epochs. We observe that repeated exposure to high-quality (HQ) data initially improves performance but eventually leads to overfitting. By quantifying memorization through loss retention and downstream evaluation scores, we show that these signals can inform the design of adaptive training schedules, enabling safe repetition of HQ data beyond the typical four-epoch limit without incurring overfitting. Our analysis offers actionable insights for \textit{memorization-aware data reuse}, laying the foundation for training LLMs \textit{smarter rather than longer} with limited high-quality data. The main contributions of this paper are three-fold:

\begin{itemize}
    \item We hypothesize the existence of a \textit{Memorization Window} in large language models and validate it through both explicit and implicit measurements across multiple metrics.

    \item We demonstrate the correlation between the \textit{Memorization Window} and the number of unique training tokens, and present preliminary results showing its impact on model performance and data scheduling strategies.

    \item We initiate a broader discussion on model memorization and its interaction with other factors influencing training efficiency, highlighting new research opportunities toward smarter and more efficient LLM training.
\end{itemize}

\section{Related Work}
\subsection{Multi-epoch training with data repetition}

\textbf{From single-epoch training to inevitable reuse.}
Under the guidance of the Chinchilla scaling law~\citep{hoffmann2022Chinchilla}, early LLMs were predominantly trained for a single epoch over the pretraining corpus. Studies from this period largely cautioned against data reuse, reporting that repeated tokens could induce memorization and harm generalization~\citep{hernandez2022scaling, lee2022deduplicating}. These findings motivated aggressive deduplication pipelines and the view that unique tokens should be prioritized over repeated exposure.
However, the landscape has shifted as models continue to scale while the supply of high-quality human-generated data becomes increasingly limited. The central question has therefore evolved from \emph{whether} to reuse data to \emph{how} it should be reused effectively.

\textbf{Empirical studies on repetition limits.}
Several works have begun to examine multi-epoch training more systematically. The scaling analysis of~\citep{taylor2022galactica, muennighoff2023scaling} shows that naively repeating data yields diminishing returns beyond roughly four epochs, establishing a widely cited practical limit. Similar observations are reported in~\citep{xue2023repeat}, which highlights the interaction between unique token count, dataset quality, and degradation under token scarcity.~\citep{parmar2024reuse} further demonstrates diminishing marginal gains from continued pretraining and explores selecting reusable subsets rather than repeating the full corpus. Complementary theory from~\citep{yan2025larger} suggests that the amount of unique data fundamentally governs how many repetitions remain beneficial and larger datasets can be
repeated more. Recently, Phi-4 \citep{abdin2024phi} reports gains when training up to 12 epochs on synthetic data, contrasting with the four-epoch heuristic.

\textbf{Data curation and controlled duplication.}
Parallel to these studies, data-centric efforts indicate that repetition can be beneficial when applied selectively. RefinedWeb \citep{penedo2023refinedweb} and FineWeb \citep{penedo2024fineweb} show that controlled duplication of curated web sub-sources can improve performance, and OLMo-3~\citep{olmo2025olmo} explicitly caps domain repetition to 4–7 passes. These results suggest that the restriction to single-epoch training is primarily a property of massive web corpora, and may not apply to carefully curated or high-quality subsets—precisely the regime where data scarcity is most acute.

\subsection{Memorization in Large Language Models}
Beyond a certain number of repetitions, models tend to memorize training data and overfit, degrading generalization. SmolLM2 \citep{allal2025smollm2} reports an early plateau when mathematics data is repeated for around ten epochs, illustrating the tension between capability gains and memorization.
Memorization itself has been characterized in multiple ways. \citep{li2025find} identify distinct phases of memorization, generalization, and grokking through validation loss dynamics, while \citep{zucchet2025emergence} shows that repetition can accelerate the emergence of capabilities in grokking-like regimes. From a distributional perspective, \citep{wang2024generalization} defines \emph{distributional memorization} as the correlation between model outputs and the pretraining data distribution. An information-theory view is provided by \citep{morris2025much}, which distinguishes intended and unintended memorization and offers statistical tools to quantify them.
Together, these works highlight that memorization is neither purely harmful nor fully understood, and that existing multi-epoch strategies lack mechanisms to monitor or exploit memorization signals—motivating our investigation of memorization-aware data reuse.

\section{Memorization and the Memorization Window}
\label{sec:memorization}

We formally define \textit{memorization} and the \textit{memorization window} to characterize when data reuse remains beneficial during LLM pretraining. Our key insight is that the interval between repeated exposures to the same sample—measured in training tokens—determines whether reuse aids learning or causes overfitting.

\subsection{Measuring Memorization}

Consider a training corpus $D_{\mathrm{train}} = (x_1, x_2, \ldots, x_n)$ processed sequentially. When the model first encounters sample $x_i$ at token position $t_i$, it incurs a certain loss. After continued training on subsequent data, the loss on $x_i$ reveals whether the model retained or forgot this sample.

\begin{definition}[Loss Retention Gap]
\label{def:rollback}
Let $\ell_i(\theta_{t_i})$ denote the loss on sample $x_i$ when first encountered at token position $t_i$, and let $\ell_i(\theta_{t_i+\tau})$ denote the loss on $x_i$ after $\tau$ additional tokens of continued training (\textbf{excluding} $x_i$). The \textbf{loss retention gap} is:
\begin{equation}
    \Delta_i(\tau) = \ell_i(\theta_{t_i})- \ell_i(\theta_{t_i+\tau}),
\end{equation}
where $\ell_i(\theta)$ is the per-sample negative log-likelihood:
\begin{equation}
    \ell_i(\theta) = -\frac{1}{|x_i|} \sum_{j=1}^{|x_i|} \log p_\theta(x_{i,j} \mid x_{i,<j}).
\end{equation}
\label{def:loss_gap}
\end{definition}

A large $\Delta_i(\tau)$ indicates the model retains information about $x_i$ after $\tau$ tokens of continual training; a small gap indicates forgetting.

\begin{definition}[Sample-level Memorization]
\label{def:memorization}
Sample $x_i$ is \textbf{memorized} at horizon $\tau_{\max}$ if the loss retention gap remains large:
\begin{equation}
    x_i \text{ is memorized} \iff \Delta_i(\tau) \geq \epsilon, \quad \forall \, \tau \in (0, \tau_{\max}],
\end{equation}
where $\epsilon > 0$ is a threshold.
Equivalently, $x_i$ is memorized if its loss remains significantly lower than when first encountered, indicating the model still retains the learned information.
\end{definition}

\subsection{The Memorization Window}

The memorization window characterizes the token interval over which data reuse remains beneficial. We define it from two complementary perspectives—retention and generalization—that correspond to our experimental measurements.

\begin{definition}[Memorization Window]
\label{def:mem_window}
We define the memorization window from two complementary perspectives:

\textbf{(i) Retention-based.} The \textbf{retention window} $\tau_{\mathrm{ret}}$ is the maximum token interval over which a sample remains memorized:
\begin{equation}
    \tau_{\mathrm{ret}} = \sup \left\{ \tau : \Delta(\tau) \geq \epsilon \right\},
\end{equation}
where $\Delta(\tau)$ is the loss retention gap (Definition~\ref{def:loss_gap}) after $\tau$ tokens of continued training. For $\tau > \tau_{\mathrm{ret}}$, the gap falls below $\epsilon$, indicating forgetting.

\textbf{(ii) Generalization-based.} The \textbf{generalization window} $\tau_{\mathrm{gen}}$ is the training horizon at which downstream performance peaks:
\begin{equation}
    \tau_{\mathrm{gen}} = \arg\max_{\tau \in (0, T]} \, \mathcal{G}(\theta_{\tau}),
\end{equation}
where $T$ is the total training budget and $\mathcal{G}(\theta_\tau)$ denotes generalization performance after $\tau$ tokens of training. For $\tau > \tau_{\mathrm{gen}}$, continued training leads to overfitting.

\textbf{(iii) Effective memorization window.} The \textbf{memorization window} $\tau^*$ is the safe reuse interval:
\begin{equation}
    \tau^* = \min(\tau_{\mathrm{ret}}, \tau_{\mathrm{gen}}).
\end{equation}
\end{definition}

Intuitively, $\tau_{\mathrm{ret}}$ marks when the model \textit{forgets}, while $\tau_{\mathrm{gen}}$ marks when \textit{overfitting} begins. The effective window $\tau^*$ is constrained by whichever limit is reached first.

\subsection{Implications for Multi-Epoch Training}

We consider two approaches to multi-epoch data scheduling:

\paragraph{Global shuffling (Traditional).}
The standard approach shuffles all data globally across epochs. Given a corpus $D_{\mathrm{train}}$ repeated for $E$ epochs, all $E \times |D_{\mathrm{train|}|}|$ samples are randomly permuted before training. This treats each sample independently, ignoring when it was last seen. The interval between consecutive exposures to the same sample $x_i$ varies randomly from 1 to $E \times |D_{\mathrm{train}}|$ tokens, providing no control over memorization dynamics.

\paragraph{Spaced repetition (Memorization-aware).}
We propose scheduling data reuse based on the memorization window. Consider a training corpus $D_{\mathrm{train}} = (x_1, x_2, \ldots, x_n)$ comprising $N$ tokens, processed sequentially and repeated over multiple epochs in the same order. Each sample $x_i$ is re-encountered after exactly $N$ tokens—a fixed, controllable interval. The relationship between epoch length $N$ and memorization window $\tau^*$ determines the effectiveness of spaced repetition:

\textbf{Case 1: $N < \tau^*$ (Over-memorization risk).}
Samples are re-exposed \textit{before} being forgotten. Since the model still retains information about $x_i$, the repeated exposure provides redundant learning signal. Continued reuse in this regime risks over-memorization and eventual overfitting.

\textbf{Case 2: $N \approx \tau^*$ (Optimal reuse).}
Samples are re-exposed \textit{just as} they begin to be forgotten. This maximizes the utility of each repetition: the model refreshes decaying information precisely when needed, balancing retention and learning efficiency.

\textbf{Case 3: $N > \tau^*$ (Forgetting before reuse).}
Samples are partially forgotten before re-exposure, requiring the model to re-learn information about $x_i$. While this regime avoids over-memorization, its impact on final performance relative to $N \approx \tau^*$ remains an empirical question.

\textbf{Epoch budget constraint.}
Even when $N \geq \tau^*$, excessive repetition eventually causes overfitting due to cumulative over-exposure. We define the \textit{safe training regime} as:
\begin{equation}
    N \geq \tau^* \quad \text{and} \quad E \leq E_{\max},
\end{equation}
where $E$ is the number of epochs and $E_{\max}$ is the maximum epochs before generalization degrades:
\begin{equation}
    E_{\max} = \arg\max_{E \in \{1, \ldots, E_{\mathrm{budget}}\}} \, \mathcal{G}(\theta^{(E)}),
\end{equation}
with $\theta^{(E)}$ denoting the checkpoint after $E$ epochs, $E_{\mathrm{budget|}|}$ the maximum epochs considered and $\mathcal{G}(\cdot)$ the downstream generalization performance.

\textbf{Summary.}
Our framework yields a principled criterion for multi-epoch training: \textit{the epoch length $N$ should satisfy $N \geq \tau^*$ to avoid over-memorization, and the total number of epochs $E$ should not exceed $E_{\max}$}. Together, these constraints define the safe reuse budget for high-quality data. We empirically validate these predictions in Section~\ref{sec:experiments}.

\textbf{Remark.}
Our definition of memorization is \textit{functional} rather than \textit{extractive}. Extractive memorization~\citep{carlini2022quantifying} concerns whether a model can verbatim reproduce training data—a privacy concern. Functional memorization concerns whether a model retains learned information, measured by loss stability under continued training. This aligns with our goal of informing data scheduling, where the key question is whether re-exposure provides learning signal.

\section{Experiments}
\label{sec:experiments}

We validate the memorization window framework through two experiments: (i)~measuring the retention window $\tau_{\mathrm{ret}}$ via rollback loss, and (ii)~measuring the generalization window $\tau_{\mathrm{gen}}$ via downstream evaluation. We aim to answer: Do $\tau_{\mathrm{ret}}$ and $\tau_{\mathrm{gen}}$ exist? How does epoch size $N$ affect performance? What role does LQ data play beyond spacing?

\subsection{Retention Window}
\label{sec:retention_results}
We measure the retention window $\tau_{\mathrm{ret}}$ by evaluating how quickly the model forgets previously seen data.

\begin{figure}[t]
\centering
\includegraphics[width=\linewidth]{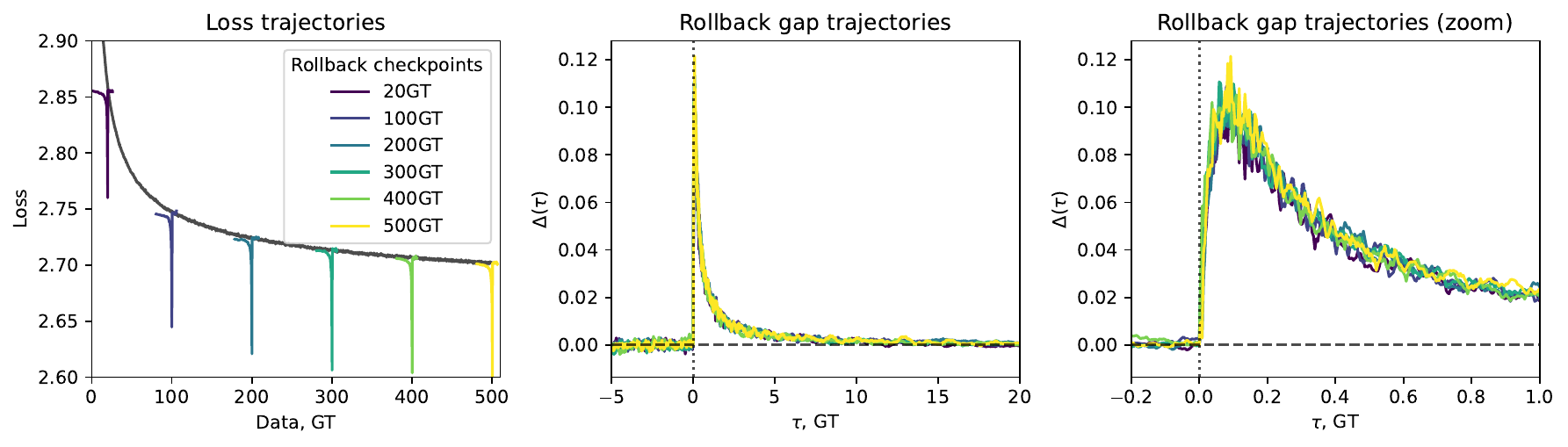}
\caption{\textit{(Left)} The loss trajectories for the original training without repetitions and for the rollbacks at the specified checkpoints from 20 to 500 Billion tokens. \textit{(Middle and right)} Rollback gap $\Delta^R(\tau)$ vs.\ data age $\tau$ (in Billion tokens).}
\label{fig:rollback}
\end{figure}

\paragraph{Setup.}
Definition~\ref{def:rollback} measures forgetting by comparing current loss to the loss when data was originally trained. In practice, tracking per-sample losses throughout training is computationally prohibitive. We adopt a simpler \textit{rollback} protocol to estimate original memorization profile $\Delta(\tau)$. Specifically, we fix model checkpoint $\theta_t$, and compute the loss of this checkpoint on the past and future tokens relative to checkpoint step $t$. Formally, we compute:
\begin{equation}\label{eq:rollback_gap}
    \Delta^R(\tau) = \ell_{\mathrm{base}}(\theta_t) - \ell(\theta_t; x_{t-\tau}),
\end{equation}
where $\ell_{\mathrm{base}}(\theta_t)$ is the estimation of the population loss of checkpoint $\theta_t$ via averaging the loss on a large enough collection of unseen token. By comparing \eqref{eq:rollback_gap} with definition \eqref{def:rollback} we see that $\Delta^R(\tau)$ reasonably approximates $\Delta(\tau)$ up to slight differences.
\begin{itemize}
    \item We can think of the retention gap $\Delta(\tau)$ as consisting of two components: the memorization of sample $x_i$ after training for $\tau$ tokens; and overall improvement of model capability from $\theta_{t_i}$ to $\theta_{t_i+\tau}$ that uniformly decreases the loss on all samples, including $x_i$. Similarly, the rollback gap $\Delta^R(\tau)$ consist of the same memorization and model evolution components, but the evolution component is different as $\tau$ is varied. Instead of the change in model capabilities when the loss is measured, rollback gap is potentially affected by a change of capabilities when the sample $x_{t-\tau}$ was memorized by the model in the state $\theta_{t-\tau}$.     
    \item Rollback gap is a noisy estimation of underlying population characteristic due to variance of per-sample loss values $\ell(\theta;x)$ with respect to $x$. We have already mentioned that $\ell_{\mathrm{base}}(\theta_t)$ is a noisy estimation of population loss. Likewise, $\ell(\theta_t; x_{t-\tau})$ can be treated as a stochastic process with respect to $\tau$, and we want to access the mean of this process. This can be done via smoothing, but choosing the smoothing window presents the classical bias-variance tradeoff.   
\end{itemize}




\paragraph{Results.}
We train a 100M model on Fineweb~\citep{penedo2024fineweb} without repetitions, and then perform rollbacks at different positions during training. The training is done over 500GT with constant learning rate and the other optimization hyperparameters. The results are depicted on figure \ref{fig:rollback}. We make two observations:
\begin{itemize}
    \item The rollback gap $\Delta^R(\tau)$ peaks at around $\tau=0.15$GT, and then quickly decays to $0$. While we don't fix the threshold $\eps$ at this moment, we see that the gap decays substantially between 1 and 5 billion tokens, putting the retention window in the range $\tau_\mathrm{ret}\in (1 \mathrm{GT}, 5\mathrm{GT})$.
    \item Perhaps counterintuitively, the rollbacks trajectories fully overlap even though the respective checkpoints span large range of training stage from 20 to 500 GT. From this observation, we cautiously conclude that the degree of memorization of individual tokens in independent from model capabilities if model architecture and optimizer hyperparameter are fixed.
\end{itemize}



\subsection{Generalization Window}
\label{sec:gen_results}
\textbf{Setup.}
We train a 100M parameter decoder-only Transformer from scratch. We use OpenMathInstruct2~\citep{toshniwal2024openmathinstruct} as high-quality (HQ) data (samping \texttt{0.8B unique tokens}) and FineWeb~\citep{penedo2024fineweb} as low-quality (LQ) data. We vary epoch size $N \in \{0.8, 0.85, 0.95, 1.6, 3.2\}\text{B tokens}$ by mixing HQ with different amounts of LQ data while keeping unique HQ tokens fixed. We evaluate on MATH500~\citep{lightman2023lets} and report accuracy vs.\ cumulative HQ tokens.

We present results in three parts: (i)~the effect of epoch size on peak location, (ii)~saturation beyond the generalization window, and (iii)~the role of LQ data diversity.

\begin{figure}[t]
\centering
\begin{subfigure}[t]{0.32\linewidth}
    \includegraphics[width=\linewidth]{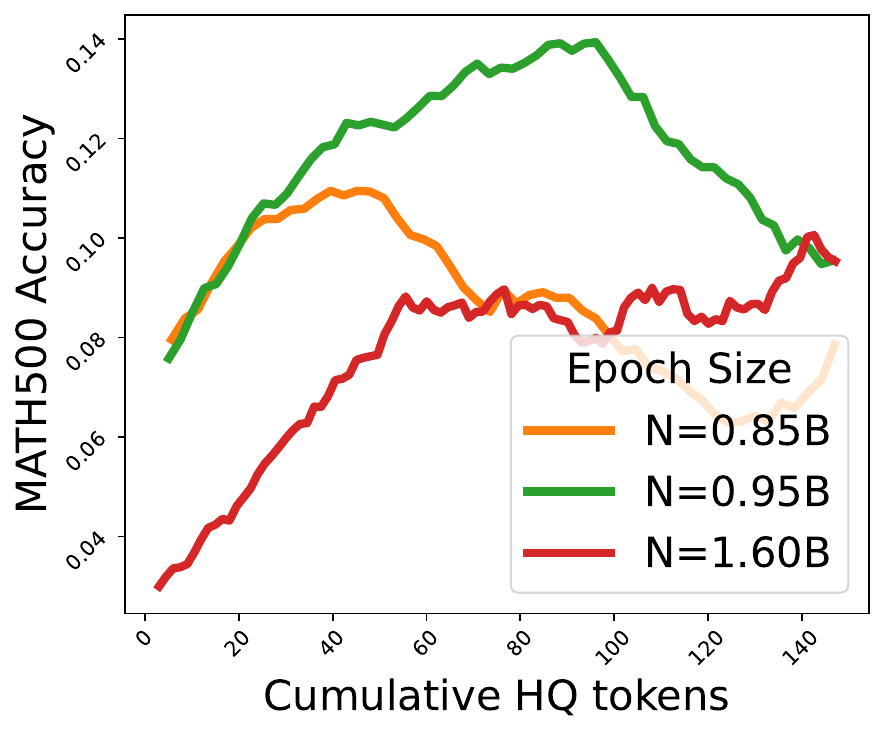}
    \caption{Epoch size effect}
    \label{fig:epoch_size}
\end{subfigure}
\hfill
\begin{subfigure}[t]{0.32\linewidth}
    \includegraphics[width=\linewidth]{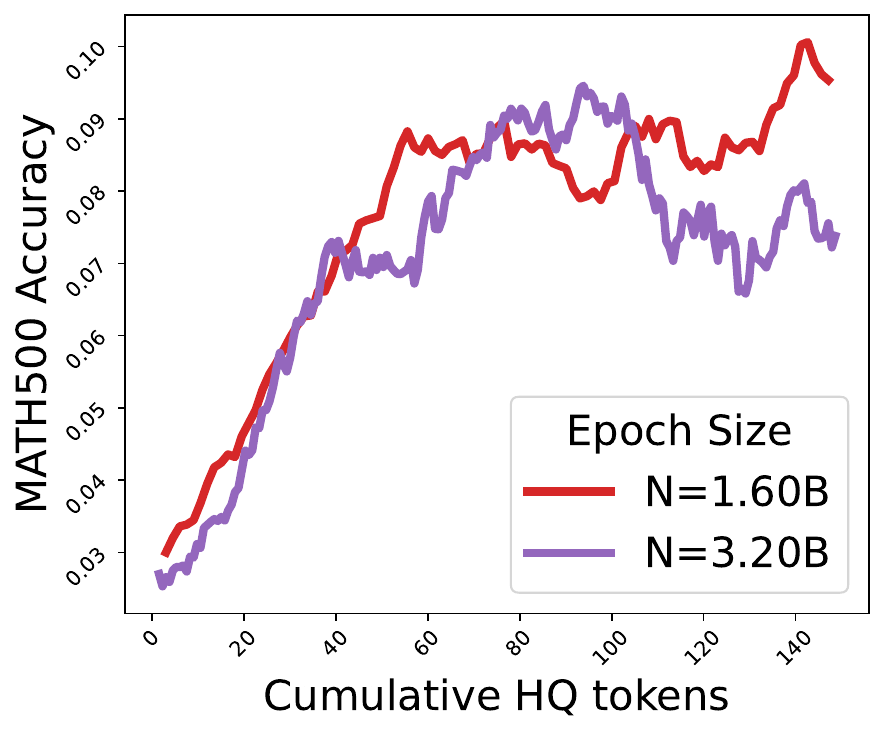}
    \caption{Saturation regime}
    \label{fig:saturation}
\end{subfigure}
\hfill
\begin{subfigure}[t]{0.32\linewidth}
    \includegraphics[width=\linewidth]{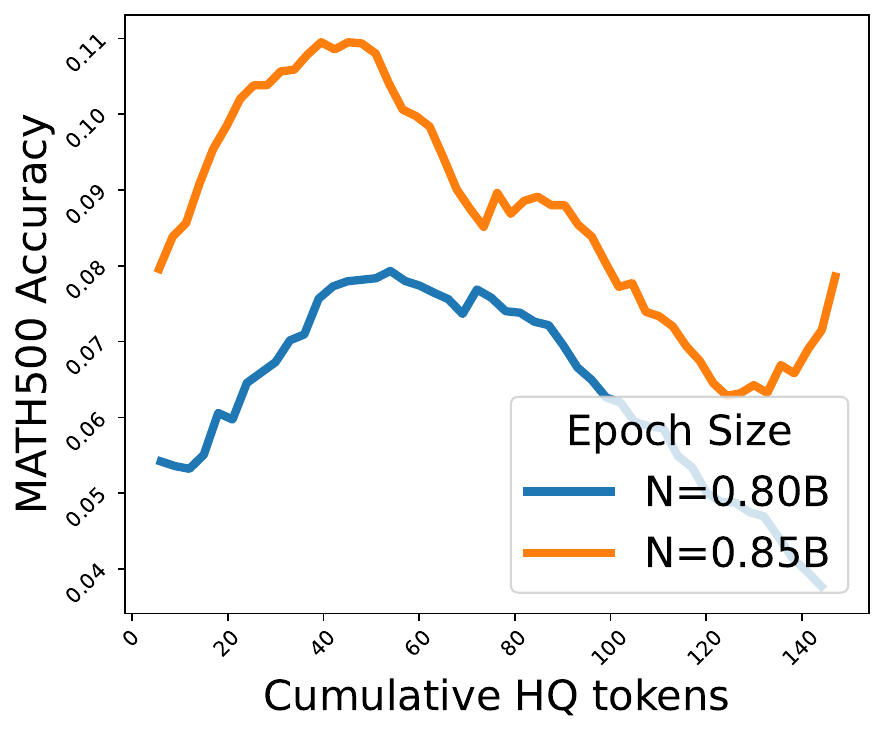}
    \caption{LQ diversity effect}
    \label{fig:lq_diversity}
\end{subfigure}
\caption{MATH500 accuracy vs.\ cumulative HQ tokens. (a)~Larger epoch size delays the peak: orange ($N=0.85\text{B}$) peaks earliest, green ($N=0.95\text{B }$) later, red ($N=1.6\text{B}$) latest among the three. (b)~Red ($N=1.6\text{B}$) and purple ($N=3.2\text{B}$) show similar behavior, both peaking at 100--150B HQ tokens—indicating saturation beyond $\tau^*$. (c)~Minimal LQ addition (6\%, orange) dramatically outperforms pure HQ (blue), demonstrating diversity matters beyond spacing.}
\label{fig:results}
\end{figure}

\textbf{Epoch size determines peak location.}
Figure~\ref{fig:epoch_size} compares configurations with $N \in \{0.85, 0.95, 1.6\}\text{B tokens}$ (orange, green, red). All exhibit rise-then-fall patterns, confirming $\tau_{\mathrm{gen}}$ exists. Larger epoch sizes delay the peak: orange peaks at $\sim$30 epochs, green at $\sim$125 epochs, and red at $\sim$175 epochs. This validates our framework—larger $N$ provides more spacing between HQ re-exposures, allowing the model to train longer before over-memorization.

\textbf{Saturation beyond the memorization window.}
Figure~\ref{fig:saturation} compares red ($N = 1.6\text{B tokens}$) and purple ($N = 3.2\text{B tokens}$). Despite a 2$\times$ difference in epoch size, both curves exhibit remarkably similar behavior—peaking at 100--150B tokens cumulative HQ tokens with comparable accuracy. This suggests \textit{saturation}: once $N$ exceeds $\tau^*$, further increases provide no additional benefit. The model already has sufficient spacing to avoid over-memorization; the limiting factor becomes the data mixture or the forgetting between exposures rather than reuse interval.

\textbf{LQ diversity is essential.}
Figure~\ref{fig:lq_diversity} compares pure HQ (blue, $N = 0.8\text{B tokens}$) with minimal LQ mixing (orange, $N = 0.85\text{B tokens}$, 6\% FineWeb). Despite nearly identical epoch sizes, orange dramatically outperforms blue. This gap cannot be explained by spacing alone—the 0.05B difference is negligible. FineWeb provides distribution diversity that regularizes learning, preventing overfitting to the narrow HQ distribution. This extends our framework:
\begin{equation}
    \text{Safe regime:} \quad N \geq \tau^*, \quad E \leq E_{\max}, \quad \rho_{\mathrm{LQ}} > 0.
\end{equation}

Table~\ref{tab:summary} summarizes our estimates from both experiments.

\begin{table}[t]
\centering
\caption{Memorization window estimates from experiments.}
\label{tab:summary}
\vspace{0.3em}
\small
\begin{tabular}{lcl}
\toprule
\textbf{Parameter} & \textbf{Estimate} & \textbf{Evidence} \\
\midrule
$\tau_{\mathrm{ret|}|}$ & 1B - 5B tokens & Retention loss (Fig.~\ref{fig:rollback}) \\
Optimal $N^*$ & $\approx 0.95\text{B tokens}$ & Highest peak (Fig.~\ref{fig:epoch_size}) \\
$\tau_{\mathrm{gen}}$ & $\geq 1.6\text{B tokens}$ & Red $\approx$ Purple (Fig.~\ref{fig:saturation}) \\
$E_{\max}$ & 125--190 epochs & $\tau_{\mathrm{gen}} / 0.8\text{B tokens}$ \\
\bottomrule
\end{tabular}
\end{table}

\textbf{Limitations.}
Our study has several limitations. First, we evaluate one domain (math), one model scale (100M), one HQ budget (0.8B tokens), and one LQ source (FineWeb)—the memorization window likely varies across these factors. Second, identifying the optimal $N^*$ by the highest peak is not fully rigorous: benchmark performance depends not only on memorization dynamics but also on the data mixture itself. For instance, smaller $N$ configurations (e.g., green) have higher HQ ratios than larger $N$ configurations (e.g., red, purple), confounding the comparison. Disentangling these effects requires further investigation. Finally, some configurations (e.g., $N=0.85B$) exhibit secondary performance rises after extended training, potentially related to grokking~\citep{li2025find}; we focus on the first peak and leave multi-phase dynamics to future work.

\section{Discussions and Perspectives}
\label{sec:discussion}
Our work introduces the memorization window as a framework for understanding data reuse in LLM training. While our experiments provide initial validation, they also open numerous research directions. We discuss the broader implications and outline key questions for future investigation.

\subsection{Implications for Data Scheduling}

\paragraph{Beyond global shuffling.}
Current multi-epoch training practices typically shuffle data globally across epochs, treating all samples uniformly regardless of when they were last seen~\citep{muennighoff2023scaling}. Our framework suggests a more principled approach: \textit{schedule data reuse based on the memorization window}. Rather than random shuffling, samples should be re-exposed when they approach the forgetting boundary ($\tau \approx \tau^*$), maximizing the learning signal from each repetition.

\paragraph{Deterministic data loading.}
Recent work on deterministic data loading~\citep{zuo2025falcon} demonstrates the benefits of sequential sample processing: reproducible training, flexible checkpointing, and dynamic mixture updates. Our framework complements this approach by providing a principled criterion for \textit{when} to revisit data sources. Specifically, if a data source has epoch size $N < \tau^*$, the scheduler should interleave other data to increase the effective reuse interval, preventing over-memorization.

\paragraph{Adaptive scheduling.}
An exciting direction is \textit{memorization-aware adaptive scheduling}: monitoring rollback loss during training and dynamically adjusting data replay based on per-sample or per-source memorization signals. Samples approaching the forgetting boundary could be prioritized for re-exposure, while well-memorized samples could be delayed. This would enable efficient use of limited high-quality data without manual tuning of epoch sizes.

\subsection{Scaling Properties of the Memorization Window}

A central open question is how the memorization window scales with model and data characteristics.

\paragraph{Model size.}
Larger models have greater capacity to retain information. We hypothesize that $\tau_{\mathrm{ret|}|}$ increases with model size—larger models forget more slowly. Counterintuitively, this implies that larger models are \textit{more susceptible} to over-memorization: since they retain information longer, repeated exposure within the same interval provides redundant signal. To avoid overfitting, larger models may require \textit{larger epoch sizes} $N$ (more spacing between repetitions) or \textit{fewer total epochs} $E$. Conversely, smaller models forget faster and may tolerate more frequent data reuse without over-memorization. This has practical implications: scaling up model size may necessitate scaling up unique data or extending reuse intervals accordingly.

\paragraph{Model architecture.}
Different architectures may exhibit different memorization dynamics. Depth versus width could affect how information is stored and forgotten—deeper models may retain more through compositional representations~\citep{zuo2025falcon}. Across architecture families, standard Transformers, linear attention models, state-space models (e.g., Mamba~\citep{gu2024mamba}), and hybrid architectures~\citep{qwen_qwen3_coder_next_tech_report} likely have distinct retention properties due to their different mechanisms for information routing. Models with explicit memory or retrieval-augmented mechanisms~\citep{huang2024ultra,cheng2026conditional} may further decouple memorization from weights. Comparing $\tau_{\mathrm{ret}}$ across architectures could reveal which designs are most efficient for learning from limited data.

\paragraph{Training stage.}
Our experiments show consistent $\tau_{\mathrm{ret}}$ across early (5B tokens) and late (500B tokens) training. However, one might expect the memorization window to \textit{expand} as training progresses—more capable models may retain information longer. If confirmed, the optimal epoch size $N^*$ should increase during training: early stages could tolerate frequent reuse, while later stages require more spacing. This connects to adaptive scheduling—dynamically adjusting $N$ based on the evolving memorization window rather than using a fixed schedule throughout training.

\subsection{Data-Dependent Memorization}

The memorization window likely varies across data characteristics and sources, with implications for scheduling in complex training setups.

\paragraph{Data complexity and domain.}
Not all data is equally memorable. High-entropy, complex samples (e.g., diverse reasoning chains) may be harder to memorize than low-entropy, repetitive patterns (e.g., templated text), suggesting complex data may need more frequent re-exposure. Similarly, different domains (code, natural language, scientific text) likely exhibit different memorization patterns due to varying structure and redundancy. We evaluate only mathematical reasoning; characterizing domain-specific memorization windows is an important direction. This suggests that the memorization window is an inherent property of an LLM-data pair, not just the model alone.

\paragraph{Complex data mixtures.}
Real-world pretraining involves mixtures of diverse data sources, each potentially with its own $\tau_{\mathrm{ret}}^{(k)}$ and $\tau_{\mathrm{gen}}^{(k)}$. Optimal scheduling would respect these heterogeneous windows rather than applying a uniform epoch size. Furthermore, training on one source may affect retention of another—sources sharing common patterns, tokens, or skills may interfere or reinforce each other's memorization. Since sources also contribute differently to downstream performance, integrating memorization windows with data contribution estimation could enable mixture strategies that balance retention, interference, and contribution weighting.

\subsection{Connection to Grokking}

Our experiments reveal secondary performance rises after initial decline—a pattern reminiscent of \textit{grokking}~\citep{li2025find}, where models suddenly generalize long after memorizing training data. This raises intriguing questions:

\paragraph{Grokking as delayed generalization beyond the memorization window.}
The initial performance peak may correspond to the memorization window boundary, while the secondary rise reflects a distinct learning phase where the model discovers generalizable patterns. Understanding this multi-phase dynamics could enable training strategies that intentionally push through the initial overfitting regime to reach grokking.

\paragraph{Data repetition and grokking.}
Recent work suggests repetition can accelerate grokking~\citep{zucchet2025emergence}. Our framework provides a lens to study this: does grokking require pushing past $\tau_{\mathrm{gen}}$ with continued exposure? Characterizing the relationship between memorization window, repetition, and grokking could yield new training paradigms.

\subsection{Broader Implications}

\paragraph{Hyperparameter interactions.}
Training hyperparameters may influence memorization dynamics. Higher learning rates could accelerate both learning and forgetting, potentially shrinking the memorization window. Batch size and optimizer choice may also affect how sharply the model memorizes individual samples. However, these interactions remain speculative without controlled experiments. We use fixed hyperparameters throughout our study; systematically investigating their effect on $\tau_{\mathrm{ret}}$ and $\tau_{\mathrm{gen}}$ is an avenue for future work.

\paragraph{Efficient use of limited high-quality data.}
As high-quality human-generated data becomes increasingly scarce, efficient data reuse becomes critical. Current practice often extends training duration with repeated data, hoping more exposure improves performance—but this risks over-memorization and wasted compute. The memorization window framework provides a principled alternative: rather than training \textit{longer} with blind repetition, we can train \textit{smarter} by timing data reuse to maximize learning signal. This means reusing data when the model begins to forget ($N \approx \tau^*$), stopping before overfitting ($E \leq E_{\max}$), and allocating compute to samples that still provide learning signal rather than those already memorized.

\paragraph{Curriculum learning.}
The memorization window connects to curriculum learning~\citep{bengio2009curriculum}. Traditional curriculum learning orders samples by difficulty—easy to hard. Our framework suggests an alternative: order by memorization dynamics, presenting samples when they approach forgetting rather than when freshly memorized. This \textit{forgetting-aware curriculum} could complement existing approaches. Notably, recent pretraining practices~\citep{zuo2025falcon,team2025kimi} have moved away from staged curricula toward uniform data mixtures throughout pretraining; our framework provides a principled basis for revisiting data scheduling strategies.

\paragraph{Continual learning.}
The retention window $\tau_{\mathrm{ret}}$ directly relates to catastrophic forgetting in continual learning~\citep{parmar2024reuse}. When learning new tasks sequentially, models forget previous knowledge—the same phenomenon our rollback loss measures. Our framework could inform replay buffer strategies: rather than replaying old data at fixed intervals or random sampling, replay could be timed based on $\tau_{\mathrm{ret|}|}$—revisiting data as it approaches the forgetting boundary, preventing forgetting without over-rehearsal.

\section{Conclusion}
\label{sec:conclusion}

We introduce the \textit{Memorization Window} framework to characterize when data reuse remains beneficial during LLM pretraining. Through the loss retention gap, we formally define sample-level memorization and derived two complementary windows: the retention window $\tau_{\mathrm{ret}}$ (when forgetting begins) and the generalization window $\tau_{\mathrm{gen}}$ (when overfitting begins). The effective memorization window $\tau^* = \min(\tau_{\mathrm{ret}}, \tau_{\mathrm{gen}})$ provides a principled criterion for data reuse.

Our experiments on a 100M parameter Transformer validated the framework: larger epoch sizes delay overfitting, excessive spacing yields diminishing returns, and low-quality data provides essential diversity beyond temporal spacing. These findings yield practical guidelines for multi-epoch training—reuse data at intervals $N \geq \tau^*$ and stop before $E_{\max}$ epochs—enabling practitioners to train LLMs \textit{smarter rather than longer} with limited high-quality data.

Limitations remain: we evaluate one domain (mathematical reasoning), one model scale (100M), and limited data configurations. More comprehensive studies are needed, e.g., investigating scaling laws across model sizes, diverse domains, data mixtures, and hyperparameter interactions. Additionally, developing efficient online methods to estimate $\tau^*$ during training would enable practical adoption. We hope this framework stimulates further research toward principled data reuse in LLM pretraining.

\bibliography{iclr2026_conference}
\bibliographystyle{iclr2026_conference}


\end{document}

%% file: math_commands.tex
\usepackage{amsmath,amsfonts,bm}

\def\eqref#1{equation~\ref{#1}}

\def\1{\bm{1}}

\def\eps{{\epsilon}}

\DeclareMathAlphabet{\mathsfit}{\encodingdefault}{\sfdefault}{m}{sl}
\SetMathAlphabet{\mathsfit}{bold}{\encodingdefault}{\sfdefault}{bx}{n}



%% file: iclr2026_conference.bib
@article{hoffmann2022Chinchilla,
  title={Training compute-optimal large language models},
  author={Hoffmann, Jordan and Borgeaud, Sebastian and Mensch, Arthur and Buchatskaya, Elena and Cai, Trevor and Rutherford, Eliza and Casas, Diego de Las and Hendricks, Lisa Anne and Welbl, Johannes and Clark, Aidan and others},
  journal={arXiv preprint arXiv:2203.15556},
  year={2022}
}

@article{hernandez2022scaling,
  title={Scaling laws and interpretability of learning from repeated data},
  author={Hernandez, Danny and Brown, Tom and Conerly, Tom and DasSarma, Nova and Drain, Dawn and El-Showk, Sheer and Elhage, Nelson and Hatfield-Dodds, Zac and Henighan, Tom and Hume, Tristan and others},
  journal={arXiv preprint arXiv:2205.10487},
  year={2022}
}

@inproceedings{lee2022deduplicating,
  title={Deduplicating training data makes language models better},
  author={Lee, Katherine and Ippolito, Daphne and Nystrom, Andrew and Zhang, Chiyuan and Eck, Douglas and Callison-Burch, Chris and Carlini, Nicholas},
  booktitle={Proceedings of the 60th Annual Meeting of the Association for Computational Linguistics (Volume 1: Long Papers)},
  pages={8424--8445},
  year={2022}
}

@article{muennighoff2023scaling,
  title={Scaling data-constrained language models},
  author={Muennighoff, Niklas and Rush, Alexander and Barak, Boaz and Le Scao, Teven and Tazi, Nouamane and Piktus, Aleksandra and Pyysalo, Sampo and Wolf, Thomas and Raffel, Colin A},
  journal={Advances in Neural Information Processing Systems},
  volume={36},
  pages={50358--50376},
  year={2023}
}

@article{taylor2022galactica,
  title={Galactica: A large language model for science},
  author={Taylor, Ross and Kardas, Marcin and Cucurull, Guillem and Scialom, Thomas and Hartshorn, Anthony and Saravia, Elvis and Poulton, Andrew and Kerkez, Viktor and Stojnic, Robert},
  journal={arXiv preprint arXiv:2211.09085},
  year={2022}
}

@article{xue2023repeat,
  title={To repeat or not to repeat: Insights from scaling llm under token-crisis},
  author={Xue, Fuzhao and Fu, Yao and Zhou, Wangchunshu and Zheng, Zangwei and You, Yang},
  journal={Advances in Neural Information Processing Systems},
  volume={36},
  pages={59304--59322},
  year={2023}
}

@article{parmar2024reuse,
  title={Reuse, don't retrain: A recipe for continued pretraining of language models},
  author={Parmar, Jupinder and Satheesh, Sanjev and Patwary, Mostofa and Shoeybi, Mohammad and Catanzaro, Bryan},
  journal={arXiv preprint arXiv:2407.07263},
  year={2024}
}

@article{yan2025larger,
  title={Larger Datasets Can Be Repeated More: A Theoretical Analysis of Multi-Epoch Scaling in Linear Regression},
  author={Yan, Tingkai and Wen, Haodong and Li, Binghui and Luo, Kairong and Chen, Wenguang and Lyu, Kaifeng},
  journal={arXiv preprint arXiv:2511.13421},
  year={2025}
}

@article{abdin2024phi,
  title={Phi-4 technical report},
  author={Abdin, Marah and Aneja, Jyoti and Behl, Harkirat and Bubeck, S{\'e}bastien and Eldan, Ronen and Gunasekar, Suriya and Harrison, Michael and Hewett, Russell J and Javaheripi, Mojan and Kauffmann, Piero and others},
  journal={arXiv preprint arXiv:2412.08905},
  year={2024}
}

@inproceedings{lozhkovsmollm2,
  title={Smollm2: When smol goes big—data-centric training of a fully open small language model},
  author={Lozhkov, Anton and Bakouch, Elie and Blazquez, Gabriel Martin and Penedo, Guilherme and Tunstall, Lewis and Marafioti, Andr{\'e}s and Lajar{\'\i}n, Agust{\'\i}n Piqueres and Kydl{\'\i}{\v{c}}ek, Hynek and Srivastav, Vaibhav and Lochner, Joshua and others},
  booktitle={Second Conference on Language Modeling}
}

@article{li2025find,
  title={Where to find Grokking in LLM Pretraining? Monitor Memorization-to-Generalization without Test},
  author={Li, Ziyue and Fan, Chenrui and Zhou, Tianyi},
  journal={arXiv preprint arXiv:2506.21551},
  year={2025}
}

@article{zucchet2025emergence,
  title={The emergence of sparse attention: impact of data distribution and benefits of repetition},
  author={Zucchet, Nicolas and d'Angelo, Francesco and Lampinen, Andrew K and Chan, Stephanie CY},
  journal={arXiv preprint arXiv:2505.17863},
  year={2025}
}

@article{wang2024generalization,
  title={Generalization vs Memorization: Tracing Language Models' Capabilities Back to Pretraining Data},
  author={Wang, Xinyi and Antoniades, Antonis and Elazar, Yanai and Amayuelas, Alfonso and Albalak, Alon and Zhang, Kexun and Wang, William Yang},
  journal={arXiv preprint arXiv:2407.14985},
  year={2024}
}

@article{morris2025much,
  title={How much do language models memorize?},
  author={Morris, John X and Sitawarin, Chawin and Guo, Chuan and Kokhlikyan, Narine and Suh, G Edward and Rush, Alexander M and Chaudhuri, Kamalika and Mahloujifar, Saeed},
  journal={arXiv preprint arXiv:2505.24832},
  year={2025}
}

@article{penedo2024fineweb,
  title={The fineweb datasets: Decanting the web for the finest text data at scale},
  author={Penedo, Guilherme and Kydl{\'\i}{\v{c}}ek, Hynek and Lozhkov, Anton and Mitchell, Margaret and Raffel, Colin A and Von Werra, Leandro and Wolf, Thomas and others},
  journal={Advances in Neural Information Processing Systems},
  volume={37},
  pages={30811--30849},
  year={2024}
}

@article{penedo2023refinedweb,
  title={The RefinedWeb dataset for Falcon LLM: outperforming curated corpora with web data, and web data only},
  author={Penedo, Guilherme and Malartic, Quentin and Hesslow, Daniel and Cojocaru, Ruxandra and Cappelli, Alessandro and Alobeidli, Hamza and Pannier, Baptiste and Almazrouei, Ebtesam and Launay, Julien},
  journal={arXiv preprint arXiv:2306.01116},
  year={2023}
}

@article{olmo2025olmo,
  title={Olmo 3},
  author={Olmo, Team and Ettinger, Allyson and Bertsch, Amanda and Kuehl, Bailey and Graham, David and Heineman, David and Groeneveld, Dirk and Brahman, Faeze and Timbers, Finbarr and Ivison, Hamish and others},
  journal={arXiv preprint arXiv:2512.13961},
  year={2025}
}

@article{hao2025reformulation,
  title={Reformulation for Pretraining Data Augmentation},
  author={Hao, Xintong and Zhu, Ruijie and Zhang, Ge and Shen, Ke and Li, Chenggang},
  journal={arXiv preprint arXiv:2502.04235},
  year={2025}
}

@article{allal2025smollm2,
  title={SmolLM2: When Smol Goes Big--Data-Centric Training of a Small Language Model},
  author={Allal, Loubna Ben and Lozhkov, Anton and Bakouch, Elie and Bl{\'a}zquez, Gabriel Mart{\'\i}n and Penedo, Guilherme and Tunstall, Lewis and Marafioti, Andr{\'e}s and Kydl{\'\i}{\v{c}}ek, Hynek and Lajar{\'\i}n, Agust{\'\i}n Piqueres and Srivastav, Vaibhav and others},
  journal={arXiv preprint arXiv:2502.02737},
  year={2025}
}

@inproceedings{carlini2022quantifying,
  title={Quantifying memorization across neural language models},
  author={Carlini, Nicholas and Ippolito, Daphne and Jagielski, Matthew and Lee, Katherine and Tramer, Florian and Zhang, Chiyuan},
  booktitle={The Eleventh International Conference on Learning Representations},
  year={2022}
}

@article{toshniwal2024openmathinstruct,
  title={Openmathinstruct-2: Accelerating ai for math with massive open-source instruction data},
  author={Toshniwal, Shubham and Du, Wei and Moshkov, Ivan and Kisacanin, Branislav and Ayrapetyan, Alexan and Gitman, Igor},
  journal={arXiv preprint arXiv:2410.01560},
  year={2024}
}

@article{lightman2023lets,
      title={Let's Verify Step by Step}, 
      author={Lightman, Hunter and Kosaraju, Vineet and Burda, Yura and Edwards, Harri and Baker, Bowen and Lee, Teddy and Leike, Jan and Schulman, John and Sutskever, Ilya and Cobbe, Karl},
      journal={arXiv preprint arXiv:2305.20050},
      year={2023}
}

@article{zuo2025falcon,
  title={Falcon-h1: A family of hybrid-head language models redefining efficiency and performance},
  author={Zuo, Jingwei and Velikanov, Maksim and Chahed, Ilyas and Belkada, Younes and Rhayem, Dhia Eddine and Kunsch, Guillaume and Hacid, Hakim and Yous, Hamza and Farhat, Brahim and Khadraoui, Ibrahim and others},
  journal={arXiv preprint arXiv:2507.22448},
  year={2025}
}

@inproceedings{gu2024mamba,
  title={Mamba: Linear-time sequence modeling with selective state spaces},
  author={Gu, Albert and Dao, Tri},
  booktitle={First conference on language modeling},
  year={2024}
}

@techreport{qwen_qwen3_coder_next_tech_report,
  title        = {Qwen3-Coder-Next Technical Report},
  author       = {{Qwen Team}},
  url          = {https://github.com/QwenLM/Qwen3-Coder/blob/main/qwen3_coder_next_tech_report.pdf},
  note         = {Accessed: 2026-02-03}
}

@article{huang2024ultra,
  title={Ultra-sparse memory network},
  author={Huang, Zihao and Min, Qiyang and Huang, Hongzhi and Zhu, Defa and Zeng, Yutao and Guo, Ran and Zhou, Xun},
  journal={arXiv preprint arXiv:2411.12364},
  year={2024}
}

@article{cheng2026conditional,
  title={Conditional Memory via Scalable Lookup: A New Axis of Sparsity for Large Language Models},
  author={Cheng, Xin and Zeng, Wangding and Dai, Damai and Chen, Qinyu and Wang, Bingxuan and Xie, Zhenda and Huang, Kezhao and Yu, Xingkai and Hao, Zhewen and Li, Yukun and others},
  journal={arXiv preprint arXiv:2601.07372},
  year={2026}
}

@inproceedings{bengio2009curriculum,
  title={Curriculum learning},
  author={Bengio, Yoshua and Louradour, J{\'e}r{\^o}me and Collobert, Ronan and Weston, Jason},
  booktitle={Proceedings of the 26th annual international conference on machine learning},
  pages={41--48},
  year={2009}
}

@article{team2025kimi,
  title={Kimi k2: Open agentic intelligence},
  author={Team, Kimi and Bai, Yifan and Bao, Yiping and Chen, Guanduo and Chen, Jiahao and Chen, Ningxin and Chen, Ruijue and Chen, Yanru and Chen, Yuankun and Chen, Yutian and others},
  journal={arXiv preprint arXiv:2507.20534},
  year={2025}
}
